\documentclass[conference]{IEEEtran}
\IEEEoverridecommandlockouts
\usepackage{cite}
\usepackage{amsmath,amssymb,amsfonts}
\usepackage{algorithmic}
\usepackage{graphicx}
\usepackage{textcomp}
\usepackage{color,soul}
\usepackage{xcolor}
\usepackage{amsmath}
\usepackage{fancyhdr}


\def\BibTeX{{\rm B\kern-.05em{\sc i\kern-.025em b}\kern-.08em
    T\kern-.1667em\lower.7ex\hbox{E}\kern-.125emX}}

\title{Perspective Transformation Layer
\thanks{
This paper has been accepted for publication by the 2022 International Conference on Computational Science \& Computational Intelligence (CSCI'22), Research Track on Signal \& Image Processing, Computer Vision \& Pattern Recognition.
}
}

\author{
    \IEEEauthorblockN{
    Nishan Khatri\IEEEauthorrefmark{1}, 
    Agnibh Dasgupta\IEEEauthorrefmark{1}, 
    Yucong Shen\IEEEauthorrefmark{2}, 
    Xin Zhong\IEEEauthorrefmark{1},
    Frank Y. Shih\IEEEauthorrefmark{2}}\
    
    \smallskip
    \IEEEauthorblockA{\IEEEauthorrefmark{1}
    Department of Computer Science, University of Nebraska Omaha, Omaha, NE, USA
    \\ 
    \{nkhatri, adasgupta, xzhong\}@unomaha.edu}
    \smallskip
    \IEEEauthorblockA{\IEEEauthorrefmark{2}
    Department of Computer Science, New Jersey Institute of Technology, Newark, NJ, USA
    \\ 
    \{ys496, shih\}@njit.edu}
}

\begin{document}

\maketitle

\begin{abstract}
Incorporating geometric transformations that reflect the relative position changes between an observer and an object into computer vision and deep learning models has attracted much attention in recent years. 
However, the existing proposals mainly focus on the affine transformation that is insufficient to reflect such geometric position changes. 
Furthermore, current solutions often apply a neural network module to learn a single transformation matrix, which not only ignores the importance of multi-view analysis but also includes extra training parameters from the module apart from the transformation matrix parameters that increase the model complexity. 
In this paper, a perspective transformation layer is proposed in the context of deep learning. 
The proposed layer can learn homography, therefore reflecting the geometric positions between observers and objects.
In addition, by directly training its transformation matrices, a single proposed layer can learn an adjustable number of multiple viewpoints without considering module parameters. 
The experiments and evaluations confirm the superiority of the proposed layer.
\end{abstract}


\vspace{-0.5em}
\section{Introduction}
\vspace{-0.5em}

%%%%%%%%%%%%%%
% \thispagestyle{fancy}
% \renewcommand{\headrulewidth}{0pt}
% \fancyhead[R]{Submission Type: Full/Regular Research Paper \\
% Research Track: CSCI-RTPC \\
% Contact Author: Xin Zhong}
% \fancyfoot{}
%%%%%%%%%%%%%%

Human vision is often insusceptible to viewpoint changes; for example, we can recognize an object regardless of the object’s presenting angle. 
In computer vision, realizing such an insusceptible model is of great interest because it can provide spatial robustness~\cite{sift} to a lot of application scenarios such as the classification and object detection~\cite{Lang2014SceneCB,YildirimS14}.  
One efficient approach to simulate the relative position changes between an observer and an object is to perform geometric transforms~\cite{wang2018geometry,krasheninnikov2012estimation}, where the geometry of an image is changed by projecting the coordinates without altering the pixel values. 
Typical efficient and linear geometric transforms include affine transformation which contains the rotation, scaling, translation, and shearing; and perspective transformation (PT) which indicates how the perception of an object changes when an observer's viewpoint varies.

Although convolutional neural networks (CNNs) have achieved state-of-the-art performance in many computer vision applications~\cite{Resnet,attention,Densenet}, due to the fragility of the neural networks under geometric distortions~\cite{papernot2016limitations,Distorted}, appropriately responding to the geometric transformations within deep learning frameworks remains a key challenge. 
Traditional CNNs try to address this geometric distortion issue through data augmentation and local pooling layers. 
Although data augmentation that adds geometric distortions into the training dataset could introduce some geometry information to a deep learning model, there is no performance guarantee for untrained transformations.
Consequently, prior knowledge in a specific problem domain is often required to determine the to-be-augmented operations, which is not always feasible. 
On the other hand, the local pooling layers (\textit{e.g}. a $2\times2$ or $3\times3$ local max pooling) can only mitigate the impact of translations after many rounds of downsampling, and translation is just one simple instance of geometric transformation. 
Hence, naive solutions are insufficient to model geometric transformations in deep learning, and researchers have been exploring different solutions.

Incorporating geometric transformations in deep learning has attracted increased attention in recent years~\cite{stn,icSTN,ddtn,PTN,ETN}. 
One most representative idea is to learn a transformation matrix (TM) by a CNN module. 
So, if the CNN module is inserted into a deep learning model and the learned TM is applied to transform some feature maps, a transform can be simultaneously trained with other layers. 
However, challenges still exist.
First, in most existing work, the learning outcomes are limited to the affine transformations which are insufficient to fully model and reflect the geometric position changes between an observer and an object in a real three-dimensional (3D) world. 
% that only include four operations (\textit{i.e.}, the rotation, scaling, translation, and shearing) while there can be a lot more possible geometries in various applications through perspective transformation. 
Furthermore, current modules often output a single TM, which only learns one possible transform. 
But, it is necessary to analyze multiple viewpoints of objects for a robust computer vision model~\cite{Hartley2004}. 
Finally, existing modular solutions often create extra to-be-trained module parameters that increase the model complexity.

\begin{figure}[h!]
    \centering
    \vspace{-1.0em}
    \includegraphics[width=0.7\linewidth]{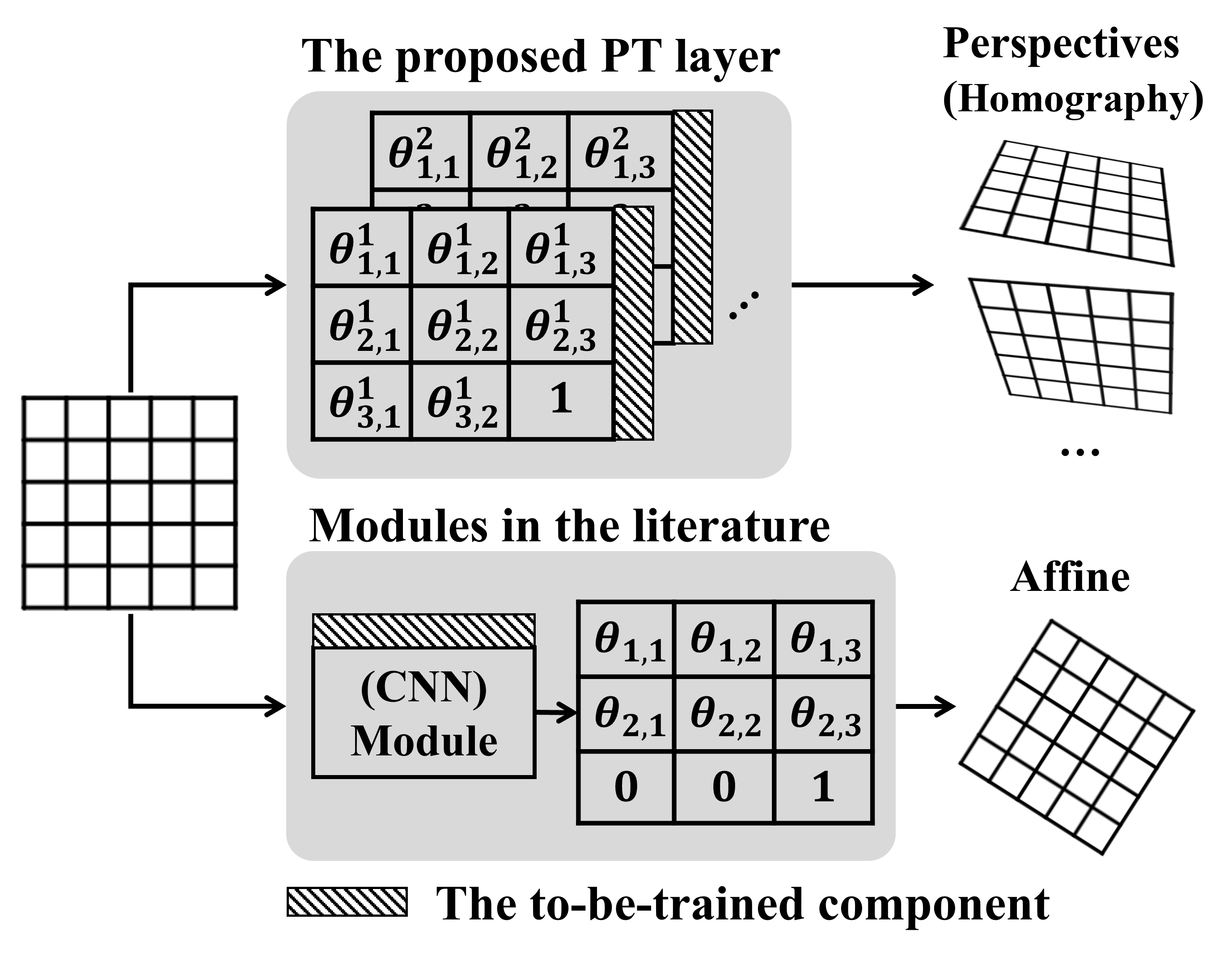}
    \vspace{-0.4cm}
    \caption{The proposed PT layer.}
    \vspace{-1.0em}
    \label{fig:1st_impression}
\end{figure}

To address the above challenges, in the context of deep neural networks, we propose a PT layer that can learn different viewpoints (see Fig.~\ref{fig:1st_impression}). 
The advantages of our design are threefold: 
(i) The PT layer learns homography so that the perception of an object from different viewpoints in a 3D world can be reflected and trained in two-dimensional (2D) images.
(ii) One single PT layer can learn multiple TMs to provide multi-viewpoint analysis to the subsequent layers; furthermore, the number of TM for a PT layer is adjustable so that the learning capacity of each PT layer can be tuned.
(iii) The proposed PT layer directly trains its TMs with gradient descent, therefore avoiding training any extra module parameters to obtain the TMs.

The remainder of this paper is organized as follows. The related work is discussed in Section~\ref{sec:related_work}. Details of the proposed PT layer is presented in Section~\ref{sec:method}. Experiments and analysis are presented in Section~\ref{sec:experiment}. Finally, conclusions are drawn in Section~\ref{sec:conclusion}.

\vspace{-1.0em}
\section{Related Work}
\label{sec:related_work}
\vspace{-0.5em}

This section briefly reviews the related work that tries to model geometric transformations in deep learning. 
Current idea is to develop a neural network module that outputs a single affine TM and to jointly train the module with other layers by transforming some feature maps with this TM.

The spatial transformer network (STN)~\cite{stn} 
% proposed by Jaderberg \textit{et al.} 
was the first effort to model the affine transformations. 
STN proposed a modular CNN block to output an affine TM so that the rotation, scaling, translation, and shearing can be applied to the module inputs.
Later, many improvements of STN were proposed. 
Incorporating a cascade of STNs, Lin \textit{et al.}~\cite{icSTN} proposed an inverse compositional STN that established a connection between the classical Lucas-Kanade algorithm~\cite{lucas1981iterative} and the STN, hence multiple affine transforms with alignments were applied to the data.
Detlefsen \textit{et al.}~\cite{ddtn} proposed a deep diffeomorphic transformer network that developed a diffeomorphic continuous piecewise affine-based transformation and created two modules that learn affine and the continuous affine respectively. 
Combining the ideas of STN and canonical coordinate representations, Esteves \textit{et al.}~\cite{PTN} proposed a polar transformer network that achieved the invariance to translations and equivariance to rotations and dilations.
% \noindent \textbf{Equivariant transformation network.} 
One recent successful improvement of STN is the equivariant transformer network (ETN)~\cite{ETN} that proposed to sequentially connect a canonical coordinates transformation and a STN module. Compared to STN, ETN improved the robustness to some continuous transformations.

There were a few proposals involving viewpoints in deep learning. 
For example, Handa \textit{et al.}~\cite{GVNN} developed a library to project 3D data into a 2D image through a single projection matrix.
Garau \textit{et al.}~\cite{garau2021deca} proposed to apply the capsule network~\cite{sabour2017dynamic} to capture the equivalent viewpoint information for human poses; 
Yan \textit{et al.}~\cite{persnet} proposed an encoder-decoder network to learn perspective information for 3D object reconstruction with only 2D supervision; 
And Yu \textit{et al.}~\cite{lanePTL} decomposed the inverse perspective mapping into multiple layers adopting encoder-decoder architecture to provide gradual viewpoint changes for lanes and road marks.
Current methods involving the modeling of perspective apply specially designed modules/networks in specific application scenarios, and hence have different aims from the proposed PT layer. 

Thus far, most existing methods use a module network to learn one affine transformation or learn viewpoint for a specific use case. 
To the best of our knowledge, our proposed PT layer is the first method that directly trains its layer TMs to learn multiple PTs for multiple possible viewpoints.

 \section{Perspective Transformation Layer}
\label{sec:method}

\begin{figure*}[h!tb]
    \centering
    \vspace{-2.0em}
    \includegraphics[width=0.75\linewidth]{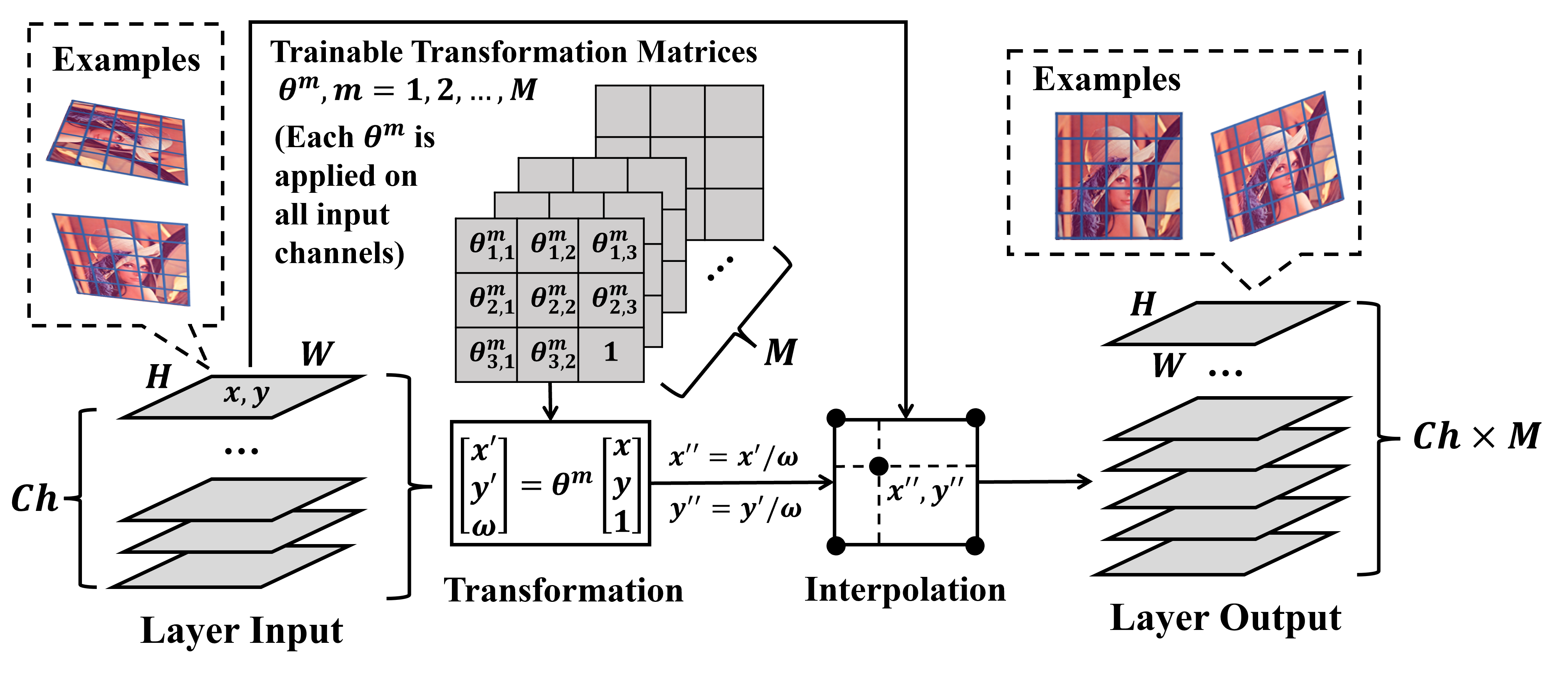}
    \vspace{-0.4cm}
    \caption{Details of the proposed PT layer.}
    \vspace{-2.0em}
    \label{fig:ptl}
\end{figure*}

Fig.~\ref{fig:ptl} presents the details of the proposed PT layer. 
In the context of deep learning, the proposed layer input is defined as images or feature maps $I\in \mathbb{R}^{N\times H\times W\times Ch}$ where $H$ denotes the height, $W$ is the width, $Ch$ represents the number of channels, and $N$ is the batch size. 
A PT layer learns $M$ TMs ($M$ is definable for an adjustable learning capacity). 
Each TM $\theta^{m} \in \mathbb{R}^{3\times 3}, m = 1,2,...,M$ transforms all the input channels. 
Thus, for each input channel, we can learn $M$ perspectives, and the layer output is defined as $O\in \mathbb{R}^{N\times H\times W \times (Ch\times M)}$ that has $Ch\times M$ channels.

\vspace{-0.5em}
\subsection{Homography Transformation Matrix}
\label{sec: transform and interpolation}

This section discusses our core learning target: the homography TM. In computer vision, under an assumption of a pinhole camera model \cite{Sturm2014}, any two images of the same planar surface in space are related by a homography. 
One major purpose in image homography is to compute a homography matrix (which estimates the parameters of the pinhole camera model). 
And once the homography matrix is obtained, we can apply it to (i) map planar surfaces in the 3D real world to the corresponding 2D images; 
and (ii) transform between different perspectives or viewpoints of the same image.

The homography matrix can be derived from a pin hole camera matrix \cite{Sturm2014} $C$ that contains the information of an internal camera matrix $IN$ and an external camera matrix $EX$:

\vspace{-0.75em}
\begin{equation}\label{eq: pin_hole_camera_matrix}
C = \\
\left[\begin{array}{llll}
C_{1,1} & C_{1,2} & C_{1,3} & C_{1,4} \\
C_{2,1} & C_{2,2} & C_{2,3} & C_{2,4} \\
C_{3,1} & C_{3,2} & C_{3,3} & C_{3,4}
\end{array}\right] 
\\
= IN \times EX.
\end{equation}

\noindent $IN$ includes information about the focal distance $f$, the scaling factors of both vertical ($x$-axis direction) and horizontal ($y$-axis direction) on an image $s_x$ and $s_y$, the translation amount $tr_x$ and $tr_y$, and a shearing factor $sh$. Using homogeneous coordinates, $IN$ can be written as:

\begin{equation}\label{eq: internal_camera}
IN=\
\left[\begin{array}{llll}
s_xf & sh & tr_x & 0 \\
0 & s_yf & tr_y & 0 \\
0 & 0 & 1 & 0 \\
\end{array}\right].
\end{equation}

\noindent And $EX$ indicates the rotation and translation of the camera frame to the world frame with its rotation parameters $r_{i,j}$ and translation amount $tr_i$:

\begin{equation}\label{eq: external_camera}
EX=\
\left[\begin{array}{llll}
r_{1,1} & r_{1,2} & r_{1,3} & tr_{1} \\
r_{2,1} & r_{2,2} & r_{2,3} & tr_{2} \\
r_{3,1} & r_{3,2} & r_{3,3} & tr_{3} \\
0 & 0 & 0 & 1 \\
\end{array}\right].
\end{equation}

\noindent $C$ maps a point $(x, y, z)$ in the 3D world space onto the image plane. 
And the homography matrix can be derived as a case of projecting a flat plane scene $(x, y, 0)$ onto the image plane.
Mathematically, 

\vspace{-1.0em}
\begin{equation}\label{eq: homography_matrix}
% \left[\begin{array}{l}
% x^{\prime} \\
% y^{\prime} \\
% \omega
% \end{array}\right]
% =
\begin{split}
\left[\begin{array}{llll}
C_{1,1} & C_{1,2} & C_{1,3} & C_{1,4} \\
C_{2,1} & C_{2,2} & C_{2,3} & C_{2,4} \\
C_{3,1} & C_{3,2} & C_{3,3} & C_{3,4}
\end{array}\right]
\times
\left[\begin{array}{l}
x \\
y \\
0 \\
1
\end{array}\right]
= \\
\left[\begin{array}{lll}
\theta_{1,1} & \theta_{1,2} & \theta_{1,3} \\
\theta_{2,1} & \theta_{2,2} & \theta_{2,3} \\
\theta_{3,1} & \theta_{3,2} & 1
\end{array}\right]
\times
\left[\begin{array}{l}
x \\
y \\
1
\end{array}\right].
% =
% \theta^{m}
% \left[\begin{array}{l}
% x \\
% y \\
% 1
% \end{array}\right]
\end{split}
\end{equation}
% \vspace{-0.75em}

We propose to train the homography matrix which has 8 degrees of freedom (\textit{i.e.}, $\theta_{1,1}$ to $\theta_{3,2}$) to learn the projection and the perspectives in the 3D world on a 2D image.
In contrast, current literature trains the affine transformation where the $\theta_{3,1}$ and $\theta_{3,2}$ in Equation~\ref{eq: homography_matrix} (\textit{i.e.}, the projection vector) are always 0. 
Consequently, the affine transformation can merely contain rotation, scaling, translation, and shearing and has to preserve parallelism while the homography does not preserve parallelism, length, or angle. 

% Hence, we propose to train the homography matrix to fully learn to project from the 3D world into a 2D image. 

% The homography TM as shown below consists of a system of homogeneous coordinates used in projective space. 
% One advantage of using homogeneous coordinates over Cartesian coordinates is that it allows for complex chaining of image transformations such as rotation and scaling using a single matrix multiplication resulting is fewer calculations.

\subsection{Transformation and Interpolation}
\label{sec: transform and interpolation}

This section describes the forward pass of the proposed PT layer: the transformation and interpolation. 
% Defining $i$-th input in the batch as $I_i \in \mathbb{R}^{H\times W \times C}$, its input pixel of $c$-th channel ($c \in [1,C]$) at location $(x, y)$ as $I_{x, y}^c$, and its corresponding $i$-th output feature map as $O_i\in \mathbb{R}^{H\times W \times C}$,
Let $(x, y)\;|\;0<=x<=W, 0<=y<=H$ denote the location of a point in one input channel, in homogeneous coordinates, the proposed PT layer transforms each point as:

\vspace{-0.75em}
\begin{equation}
\label{eq: perspective transformation}
\left[\begin{array}{l}
x^{\prime} \\
y^{\prime} \\
\omega \\
\end{array}\right] \\
=
\theta^{m} 
\times
\left[\begin{array}{l}
x \\
y \\
1 \\
\end{array}\right], \\
\end{equation}

\noindent where $\theta^{m}$ is $m$-th $3\times3$ to-be-learned TM with 8 degrees of freedom shown as $\theta_{1,1}$ to $\theta_{3,2}$ in Equation~\ref{eq: homography_matrix}, and $(x^{\prime}, y^{\prime}, \omega)$ is the transformed coordinates in which $\omega$ represents the distance between a camera and an image plane. We can convert homogeneous coordinates back to Cartesian coordinates by

\vspace{-0.75em}
\begin{equation}\label{eq: transfer_to_Cartesian}
    (x^{\prime\prime}, y^{\prime\prime}) = (x^{\prime}/\omega, y^{\prime}/\omega).
\end{equation}

\noindent Thus, giving a point $(x, y)$ on an input feature map, the PT layer transforms it to a new location $(x^{\prime\prime}, y^{\prime\prime})$ on its corresponding output perspective feature map.  

The next step is to perform interpolation to determine each new value at each new location. 
At the $ch$-th channel, denoting the value of the input at $(x, y)$ as $I^{ch}_{x,y}$ and its corresponding output perspective feature map as $O^{ch}$, we can compute the interpolation as: 

\vspace{-0.75em}
\begin{equation}\label{eq: interpolation}
    O^{ch} = \sum_{x}^H \sum_{y}^W I^{ch}_{x,y} K(x^{\prime\prime} - x, y^{\prime\prime} - y), 
\end{equation}
\vspace{-0.75em}

\noindent where $K(\cdot, \cdot)$ is a (differentiable) kernel function that can be implemented with different strategies.
For example, if we use a bilinear interpolation, the kernel function is:

\vspace{-0.75em}
\begin{equation}\label{eq:bilinear}
    K(x, y) = K_1(x)K_1(y),\ \ K_1(u) = |1- \lfloor u \rfloor |.
\end{equation}
\vspace{-1.0em}

And if we use a bicubic interpolation, $K_1$ in Equation~\ref{eq:bilinear} is:

\vspace{-0.75em}
\begin{equation}\label{eq: bicubic}
    K_1(u) = \begin{cases}
    (\alpha + 2) |u|^3 - (\alpha + 3) |u|^2 + 1,\ if\ |u| \leq 1\\
    \alpha |u|^3 - 5\alpha |u|^2 + 8 \alpha |u| - 4\alpha,\ if\ 1 < |u| < 2\\
    0,\ otherwise
    \end{cases}, 
\end{equation}
\vspace{-1.0em}

\noindent where $\alpha$ is the free parameter in a bicubic interpolation.

% By far, we have presented the the mathematical formulation of the transformation and interpolation inside the proposed PT layer.

\subsection{Back-propagation}
\label{sec: backprop}

Besides learning the homography, another advantage of the proposed PT layer is its ability to directly train its multiple TMs with gradient descent. 
So, we show that the PT layer is differentiable in this section.

The operations of the PT layer include transformation and interpolation (\textit{i.e.}, Equations~\ref{eq: perspective transformation}, \ref{eq: transfer_to_Cartesian},and \ref{eq: interpolation}).
We first show that the interpolation in Equation~\ref{eq: interpolation} is differentiable by computing the partial derivatives of the output $O^{ch}$ with respect to $I_{x, y}^{ch}$, $x^{\prime \prime}$, and $y^{\prime \prime}$:

\vspace{-0.75em}
\begin{equation}\label{eq:bp1}
    \frac{\partial O^{ch}}{\partial I_{x, y}^{ch}} = \sum_x^H \sum_y^W K(x^{\prime \prime}-y, y^{\prime \prime}-x),
\end{equation}
\vspace{-0.75em}

\vspace{-0.75em}
\begin{equation}\label{eq:bp2}
    \frac{\partial O^{ch}}{\partial x^{\prime \prime}} = \sum_x^H \sum_y^W K_1(y^{\prime \prime} - x) K_1^{\prime} (x^{\prime \prime} - y).
\end{equation}
\vspace{-0.75em}

\noindent And we can obtain $\partial O^{ch} / \partial y^{\prime \prime}$ similar to Equation~\ref{eq:bp2}.
As discussed in Equations~\ref{eq:bilinear} and~\ref{eq: bicubic}, we have implemented the bilinear and bicubic interpolation kernels.
For the bilinear kernel, Equations~\ref{eq:bp1} and~\ref{eq:bp2} can be written as:

\vspace{-0.75em}
\begin{equation}
    \frac{\partial O^{ch}}{\partial I_{x, y}^{ch}} = \sum_x^H \sum_y^W |1 - \lfloor x^{\prime \prime} -y \rfloor| |1 - \lfloor y^{\prime \prime} - x \rfloor|,
\end{equation}
\vspace{-0.75em}

\noindent and

\vspace{-0.75em}
\begin{equation}
    \frac{\partial O^{ch}}{\partial x^{\prime \prime}} = - \sum_x^H \sum_y^W |1 - \lfloor y^{\prime \prime} - x \rfloor|.
\end{equation}
\vspace{-0.75em}

\noindent And for the bicubic kernel, Equation~\ref{eq:bp1} can be written as:

\vspace{-0.75em}
\begin{equation}
    \frac{\partial O^{ch}}{\partial I_{x, y}^{ch}} = \sum_x^H \sum_y^W K_1(x^{\prime \prime} - y) K_1(y^{\prime \prime} - x),
\end{equation}
\vspace{-0.75em}

\noindent where $K_1(u)$ can be denoted by Equation~\ref{eq: bicubic}. 
In addition, the $K_1^{\prime}$ in Equation~\ref{eq:bp2} is:

\begin{equation}
    K_1^{\prime}(u) = \begin{cases}
    -3 \alpha u^2 + 10 \alpha u + 8 \alpha,\ if\ -2 < u < -1,\\
    -3 |\alpha + 2 | u^2 - 2(\alpha + 3) u,\ if\ -1 \leq u < 0,\\
    3|\alpha + 2 | u^2 - 2(\alpha+3) u,\ if\ 0 < u \leq 1,\\
    3\alpha u^2 - 10 \alpha u + 8\alpha,\ if\ 1 < u <2,\\
    0,\ otherwise.
    \end{cases}.
\end{equation}
\vspace{-0.75em}

\noindent So far, we have shown that the interpolation in Equation~\ref{eq: interpolation} is differentiable.

When performing back-propagation, the gradient flows through the interpolation (Equation~\ref{eq: interpolation}), the coordinates conversion (Equation~\ref{eq: transfer_to_Cartesian}), and the transformation (Equation~\ref{eq: perspective transformation}).
For the coordinates conversion, it directly transmits the gradient obtained by the interpolation without altering anything.
And finally, the transformation is differentiable and its derivative is $(x, y, 1)^T$.

In sum, the proposed PT layer is differentiable so that it is stackable in deep neural networks, and the TMs can be trained by gradient descent together with other model weights.
Notably, because each TM ($\theta^m$) is applied in the PT of all input channels, it affects multiple positions. 
To update the parameters in a TM, we average all the derivatives obtained from all the affected positions.

\section{Experiments and Analysis}
\label{sec:experiment}

This section experimentally analyzes quantitative and analytical evaluation of the proposed PT layer. 
Section~\ref{sec: datasets} introduces our data preparation.
Section~\ref{sec: functionality} validates the functionality of the proposed PT layer by learning known target PTs.
Section~\ref{sec: ablation} discusses the ablation study that compares the performance with and without the PT layer.
Section~\ref{sec: TMs} analyzes the influence of multiple TMs of the PT layer.
Finally, Section~\ref{sec: comparison} compares the performance of the proposed PT layer against other well-known modules. 
Besides, some additional experiments of image segmentation and potential application discussions and are in the project's Github page\footnote {https://github.com/kcnishan/Perspective\_Transformation\_Layer.git}. 

\subsection{Data Preparation}
\label{sec: datasets}

\begin{figure}[h!]
    \centering
    \vspace{-0.75em}
    \includegraphics[width=0.50\linewidth]{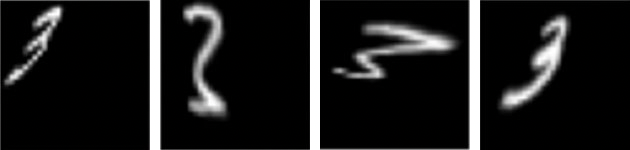}
    \vspace{-0.4em}
    \caption{A few MNIST images with random PTs.}
    \vspace{-1.0em}
    \label{fig:pers_MNIST}
\end{figure}

\noindent \textbf{Distorted MNIST.} 
The modified national institute of standards and technology database (MNIST)~\cite{mnist} is a standard dataset of handwritten digits that consists of $60,000$ training and $10,000$ testing grayscale images of size $28\times28$.
For our training dataset, 
% \hl{Nishan: Added this} we  resize each image to $32*32$, then 
we  apply random PTs on $55,000$ images and leave the remaining $5,000$ unmodified. 
Some images are left unmodified since the unmodified images can provide the original geometric information to the PT layer. 
For the testing set, we apply random PTs on all $10,000$ images. 
Figure~\ref{fig:pers_MNIST} shows a few examples of randomly transformed MNIST images.

\begin{figure}[h!]
    \centering
    \vspace{-0.75em}
    \includegraphics[width=0.50\linewidth]{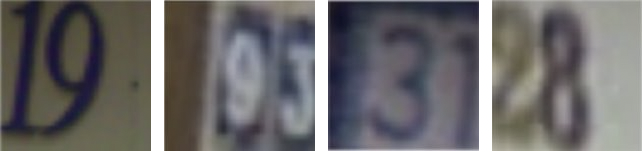}
    \vspace{-0.4em}
    \caption{A few SVHN images.}
    \vspace{-0.75em}
    \label{fig:pers_SVHN}
\end{figure}

\noindent \textbf{SVHN.} 
The street view house numbers dataset (SVHN)~\cite{svhn} is a dataset containing $73,257$ training and $26,032$ testing RBG color images of size $32\times32$ (examples in Figure~\ref{fig:pers_SVHN}). 
The SVHN can facilitate our experiments because the street number images are collected by cameras that naturally introduce various viewpoints.
In our experiments, we show that the PT layers can learn good viewpoints to help a downstream classifier that recognizes the street numbers.

\begin{figure}[h!]
    \centering
    \vspace{-1.0em}
    \includegraphics[width=0.50\linewidth]{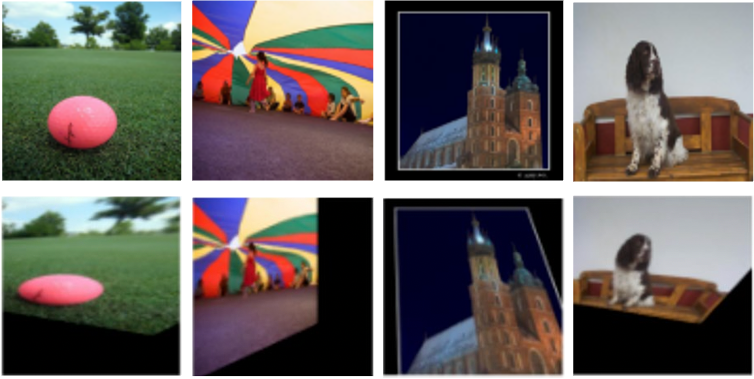}
    \vspace{-0.4cm}
    \caption{A few Imagenette examples. First row: original images; Second row: distorted images.}
    \vspace{-1.0em}
    \label{fig:pers_Imagenette}
\end{figure}

\noindent \textbf{Imagenette.} 
The Imagenette~\cite{imagenette} is a collection of 10 representative classes from the huge Imagenet dataset~\cite{deng2009imagenet}. 
The Imagenette dataset includes $9,469$ training and $3,925$ testing RGB images of various sizes. 
Out of these, we use the ones with $320\times320$ size and resized them to $224\times224$. 
We adopt this representative and small version of the Imagenet for two main reasons: 
(i) the Imagenette validates the proposed PT layer with larger ($224\times224$) images compared to the MNIST and the SVHN; 
and (ii) rather than applying all the $1,000$ classes with more than millions of images from the Imagenet, the Imagenette provides a fast reproducibility. 
In our experiments, we test the proposed PT layer on both the original and distorted Imagenette for a comprehensive performance evaluation.
In terms of the distorted Imagenette, for the training set of $9,469$ images, we apply random PTs on approximately $8,285$ images (87.5\%) and leave the remaining $1,184$ images (12.5\%) unmodified. 
For the $3,925$ testing images we use the same split of transformed and unmodified images.
Figure~\ref{fig:pers_Imagenette} shows a few examples of Imagenette images along with their distortions.

\subsection{Functionality Validation}
\label{sec: functionality}

This section validates the fundamental functionality of the proposed PT layer. 
As a layer designed to learn different viewpoints, the proposed PT layer should be able to easily rectify the inputs transformed by PTs if the original images are given as the labels.

\begin{figure}[h!]
    \centering
    \vspace{-0.75em}
    \includegraphics[width=0.85\linewidth]{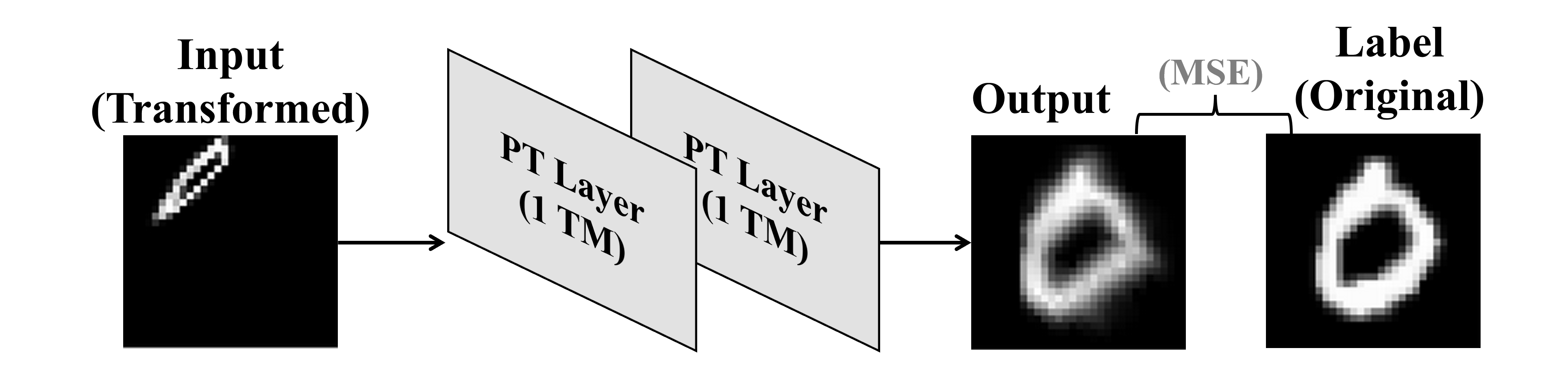}
    \vspace{-0.4cm}
    \caption{The process of the functionality validation.}
    \vspace{-0.75em}
    \label{fig: functionality}
\end{figure}

\noindent \textbf{Experiment scheme.}
As illustrated in Figure~\ref{fig: functionality}, we create a simple model with two PT layers (each layer has only 1 TM) where the model inputs are the MNIST images with PTs, and the outputs are the undistorted images.
We train this model by minimizing the mean squared error (MSE) between the model outputs and the original images.

\begin{figure}[h!]
    \centering
    \vspace{-0.25cm}
    \includegraphics[width=0.35\linewidth]{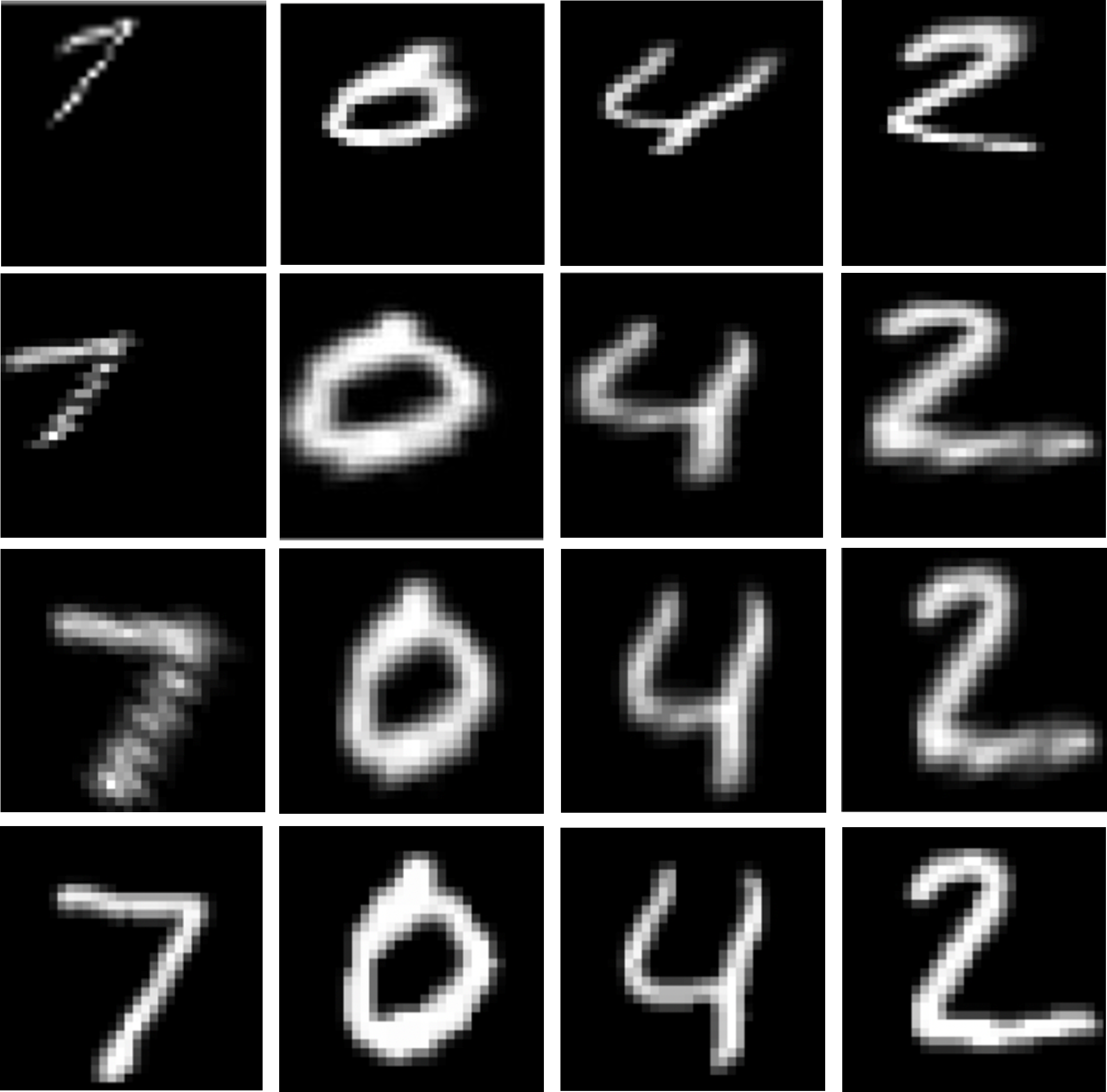}
    \vspace{-0.3cm}
    \caption{Examples from functionality validation with a two PT layer model. Top row: The input transformed MNIST images; Second row: Outputs of the first PT layer; Third row: Outputs of the second PT layer; Last row: The target original MNIST images.}
    \vspace{-1.0em}
    \label{fig:sanity2}
\end{figure}

\noindent \textbf{Results.}
This simple two-layer model can successfully learn the rectifications within a few epochs.
Although we actually find that even a single PT layer can learn this rectification, we present this two-layer model to illustrate a progressive transformation of each PT layer towards the final target (see a few examples in Figure~\ref{fig:sanity2}). 
The results of the two-layer model not only demonstrate the functionality of the PT layers but also show that the PT layers are stackable (the gradients flow through them) in a deep learning framework.

\subsection{Ablation Study}
\label{sec: ablation}

This section discusses the advantage of using the proposed PT layer. 
In addition to the study that compares the performance of a model with and without PT layer(s), we also test the effect of different inserting positions and multiple PT layers. 

\begin{figure}[h!]
    \centering
    \vspace{-1.0em}
    \includegraphics[width=0.75\linewidth]{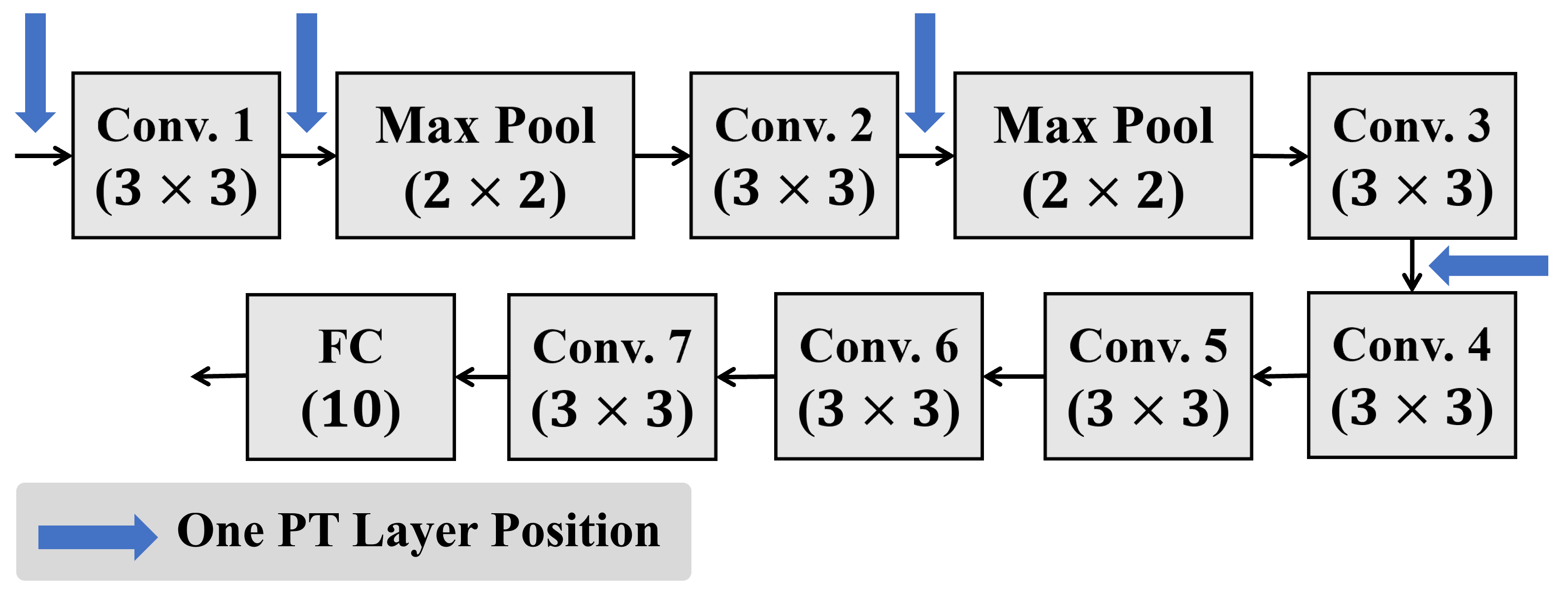}
    \vspace{-0.4cm}
    \caption{The model of our ablation study.}
    \vspace{-1.0em}
    \label{fig: ablation_model}
\end{figure}

\noindent \textbf{Experiment scheme.}
As presented in Figure~\ref{fig: ablation_model}, we adopt the CNN architecture used in the evaluation from the G-CNN~\cite{gcnn} (all convolutional layers have 32 filters) and select the places after the input, the first, the second, and the third convolutional layers as the inserting positions of a PT layer. 
To empirically test a good position for the PT layer, we insert a PT layer (with four TMs) at one position at a time. 
And to test multiple PT layers in a model, we also test the case of placing a PT layer at all the positions mentioned above.

The ablation study is performed on the distorted MNIST and the SVHN. 
We train the  models for 30 epochs  with the goal of classification using categorical cross-entropy as the loss function, which is minimized by the ADAM optimizer~\cite{kingma2014adam} at a starting learning rate of 0.0001. 
Notably, we add a $1\times1$ convolutional layer after a PT layer. 
The filter number of the convolutional layer is equal to the channel number of the input to the PT layer.
In this way, the channel shapes of the feature maps are the same before and after the PT layer.

\noindent \textbf{Results.}
Table~\ref{table:ablation_MNIST_SVHN} shows the classification testing accuracy, where one can observe that inserting the PT layer(s) improves the classification performance at each position (the baseline model is the one without any PT layers).
% Furthermore, the empirically best positions of a PT layer are the positions after the input or after the first convolution.
In addition, placing a PT layer at all the selected positions achieves the best accuracy, which indicates that increasing the number of PT layers can enhance the performance of a model. 
Given a deep learning model, the number of PT layers and the inserting positions are tunable hyperparameters.

% Ablation study table for MNIST and SVHN
\begin{table} [h!]
\centering
\begin{tabular}{ccc}
\hline
Architecture & Accuracy \% & Accuracy \%\\
(G-CNN) & (Distorte MNIST) & (SVHN)\\
\hline
% \multirow{3}{4em}{Multiple row} & cell2 & cell3 \\ 
Baseline & 95.97 & 89.56 \\ 
After input & 96.55 & 90.06 \\ 
After conv. 1 & 96.22 & 90.89 \\ 
After conv. 2 & 96.37 & 90.38 \\
After conv. 3 & 96.41 & 90.30 \\
All positions & 97.20 & 91.35 \\
\hline
\end{tabular}
\vspace{0.5em}
\caption{Classification results with and without the PT layer at different positions.}
\label{table:ablation_MNIST_SVHN}
\vspace{-2.0em}
\end{table}

\vspace{-0.5em}
\subsection{Multiple Transformation Matrices}
\label{sec: TMs}

This section presents the experiments of a PT layer with multiple TMs. 
The influence of different numbers of TMs is analyzed.

\noindent \textbf{Experiment scheme.} 
We use the well-known ResNet-18~\cite{Resnet} and the VGG-16~\cite{vgg16} (pretrained on the Imagenet dataset) as the baseline models. 
A PT layer with different numbers of TMs is inserted after the input layer in each baseline model. 
All the models are trained with the goal of classification using categorical cross-entropy as the loss, which is minimized by the ADAM optimizer at a starting learning rate of 0.0001. 
Thus, in order to analyze the influence of multiple TMs, the number of TMs is the only difference between the models' training configurations. 
Also, the PT layer in nature may favor the datasets including PT distortions. 
To exclude this possible bias, we test on both the distorted and original datasets.

\noindent \textbf{Results.}
Tables~\ref{table:TMs} and~\ref{table:TM2s} (PTL-$x$ denotes a PT layer with $x$ TMs, $x=$ 4,8,16,32 here) present the experimental results.
In most cases, because increasing the TM number enhances the learning capacity, a larger number of TMs can improve the model performance.
When the TM number is already large, keeping increasing the TM number could only bring incremental improvements. 
This is because it is the baseline model that cap the overall performance. 
There is one case (\textit{i.e.}, PTL-32 in the ResNet-18 on the original Imagenette) where keeping increasing the TM number decreases the performance.
This is because increasing the learning capacity of one of the layers cannot guarantee the improved performance of the whole model~\cite{NIPS2017_10ce03a1,novak2018x13}.
Given a deep learning model, the number of TMs is a tunable hyperparameter for the proposed PT layer.
% \hl{PTL-32 performance  is lower than other PTL models for original Imagenette on ResNet-18 architecture. One possible reason behind  this is that out of all TMs, some TM might bring extreme distortion to the original image.}

\begin{figure}[h!]
    \centering
    \vspace{-1.0em}
    \includegraphics[width=0.3\linewidth]{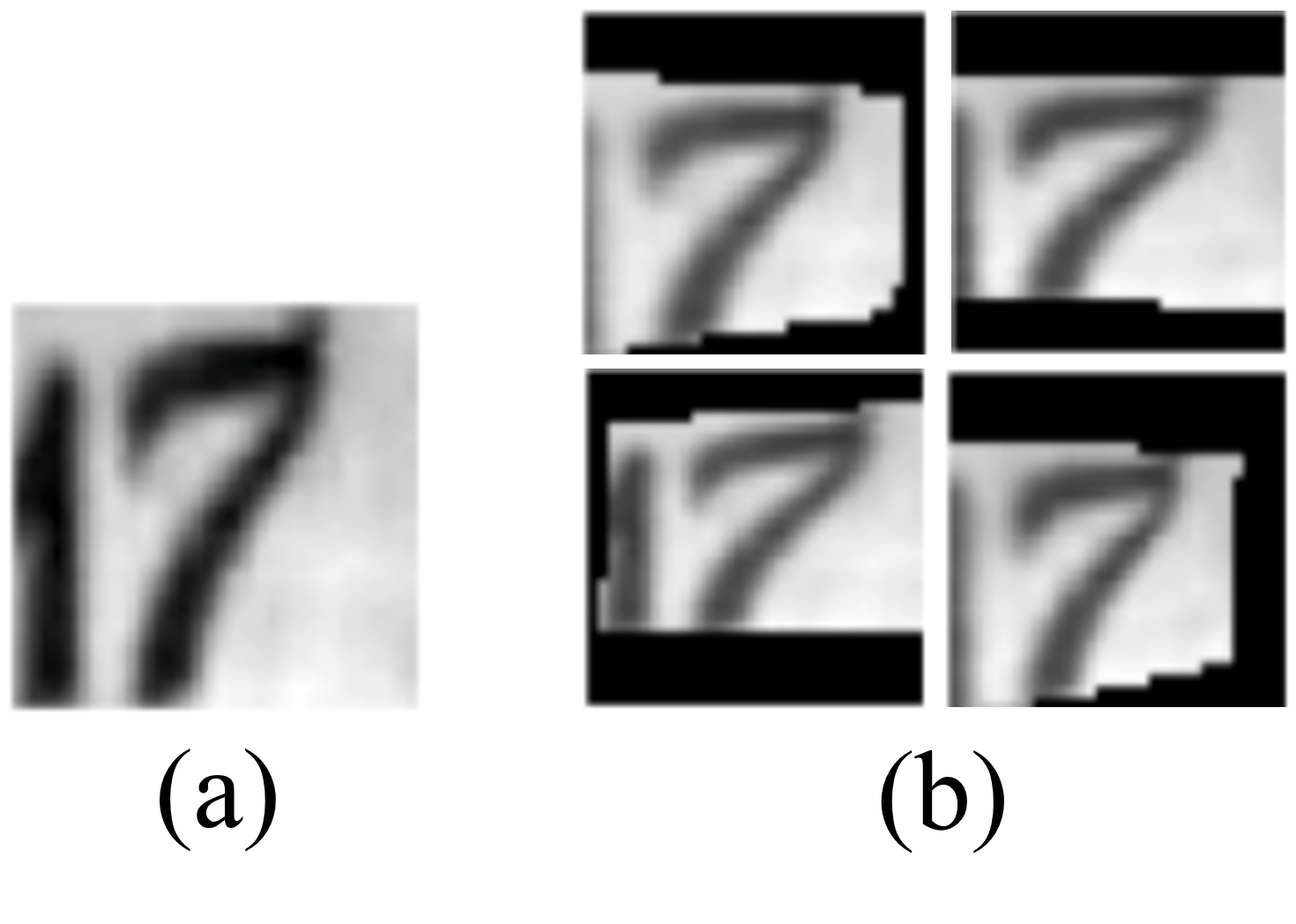}
    \vspace{-0.4cm}
    \caption{An example of learned viewpoints. (a) An input image; (b) Its various views learned by the PT layer.}
    \label{fig: SVHN_multiview}
    \vspace{-1.0em}
\end{figure}

In addition, we have validated that each trained TM in a PT layer is different so that various possible viewpoint options are learned and offered to the successive layers.
Figure~\ref{fig: SVHN_multiview} shows an example image from the SVHN dataset and a few of its learning outcomes, where we can observe different viewpoints (caused by different learned TMs).

\begin{table} [h!]
\centering
\begin{tabular}{ccccc}
\hline
Architecture & Accuracy \% & Accuracy \% \\
& (Distorted MNIST) & (Distorted Imagenette) \\
& (ResNet-18) & (VGG-16)\\
\hline
Baseline & 97.63 & 81.40 \\
PTL-4 & 97.66 & 87.62 \\ 
PTL-8 & 97.74 & 90.42 \\
PTL-16 & 97.87 & 91.31 \\
PTL-32 & 97.95 & 91.26 \\
\hline
\end{tabular}
\vspace{0.5em}
\\
% {\raggedright 
% \small
% Note: PTL-$x$ denotes a PT layer with $x$ TMs ($x=$ 4,8,16, etc). \par}
\caption{Different number of TMs on the distorted MNIST and Imagenette.}
\label{table:TMs}
\end{table}
\vspace{-1.0em}

\begin{table} [h!]
\centering
\begin{tabular}{cccc}
\hline
Architecture & Accuracy \% & Accuracy \% & Accuracy \%\\
& (Original & (Original & (SVHN)\\
& Imagenette) & Imagenette) & \\
& (VGG-16) & (ResNet-18) & (Resnet-18) \\
\hline
Baseline & 88.56 & 76.59 & 92.23 \\ 
PTL-4 & 94.09 & 81.40 & 92.49 \\ 
PTL-8 & 95.62 & 81.96 & 92.84\\ 
PTL-16 & 96.41 & 82.42 & 93.16\\
PTL-32 & 96.76 & 80.90 & 93.44\\
\hline
\end{tabular}
\vspace{0.5em}
% {\raggedright 
% \small
% Note: PTL-$x$ denotes a PT layer with $x$ TMs ($x=$ 4,8,16,32 here). 
% \par}
\caption{Different number of TMs on the original Imagenette and SVHN.}
\label{table:TM2s}
\end{table}
\vspace{-1.0em}

\subsection{Comparison}
\label{sec: comparison}

This section compares the performance of the proposed PT layer against the state-of-the-art methods on our datasets.
Specifically, to extract geometric features, the proposed PT layer, the STN module, and the ETN module are respectively inserted after the input of a baseline model, and the classification accuracies are compared.

It is worth noting that constructing an ETN module requires providing information about the order and type of distortion to its transformer layers.
In contrast, the proposed PT layer does not need any prior knowledge of the distortions. 
In this experiment, we provide the following prior knowledge to the ETN: the distortion consists of a translation, a PT on the horizontal axis, a PT on the vertical axis, and a scaling on the horizontal axis.

Another advantage of the proposed PT layer is that it directly applies gradient descent to train its TMs. 
Thus the PT layer only requires the TM parameters and does not introduce any to-be-trained module network parameters. 
Each learning homography matrix has 8 degrees of freedom, so a PT layer has TMs $\times\;8$ $\times\;Ch$ (where $Ch$ is number of channels) training parameters (\textit{e.g.}, a PT layer will have $16\times8\times1=128$ training parameters when we have 16 TMs and single channel).
In contrast, existing modularized solutions create the extra module
parameters, for example, to learn one affine matrix, a typical STN module introduces $36,000$ extra training parameters with two convolutional and two fully-connected layers.

\noindent \textbf{Experiment scheme.} 
The performance comparison experiment scheme is similar to the one described in Section~\ref{sec: TMs}. 
We use the ResNet-18 and the VGG-16 models on different distorted and original datasets and compare the performances of the PT layer, the STN, and the ETN.
The number of TMs of the PT layer is 16 or 32 in these experiments.

\begin{table} [h!]
\centering
\begin{tabular}{ccc}
\hline
Architecture & Accuracy  \% & Accuracy \%\\
& (Dostorted MNIST)  & (Distorted Imagenette)\\
& (ResNet-18)  & (VGG-16) \\
\hline
Baseline & 97.63 & 81.40\\ 
STN & 97.74 & 82.96\\ 
ETN & 97.68  & 86.60\\ 
PT layer & 97.95 & 91.31\\
\hline
\end{tabular}
\vspace{0.5em}
% {\raggedright 
% \small
% Note: PTL-$x$ denotes a PT layer with $x$ TMs ($x=$ 4,8,16,32 here). 
% \par}
\caption{Comparison of the PT layer with the STN and ETN on the distorted MNIST and Imagenette.}
\label{table:compare1}
\end{table}
\vspace{-1.0em}

%continue table
\begin{table} [h!]
\centering
\begin{tabular}{cccc}
\hline
Architecture & Accuracy \% & Accuracy \% & Accuracy \%\\
& (Original & (Original & (SVHN)\\
& Imagenette) & Imagenette) & \\
& (VGG-16) & (ResNet-18) & (Resnet-18) \\
\hline
Baseline & 88.56 & 76.59 & 92.23 \\ 
STN & 88.82 & 81.27 & 92.56 \\ 
ETN & 94.29 & 80.56 & 92.42\\ 
PT layer & 96.76 & 82.42 & 93.44\\
\hline
\end{tabular}
\vspace{0.5em}
% {\raggedright 
% \small
% Note: PTL-$x$ denotes a PT layer with $x$ TMs ($x=$ 4,8,16,32 here). 
% \par}
\caption{Comparison of the PT layer with the STN and ETN on the original Imagenette and SVHN.}
\label{table:compare2}
\end{table}
\vspace{-1.0em}

\begin{figure*}[!h]
    \centering
    \vspace{-2.0em}
    \includegraphics[width=0.7\linewidth]{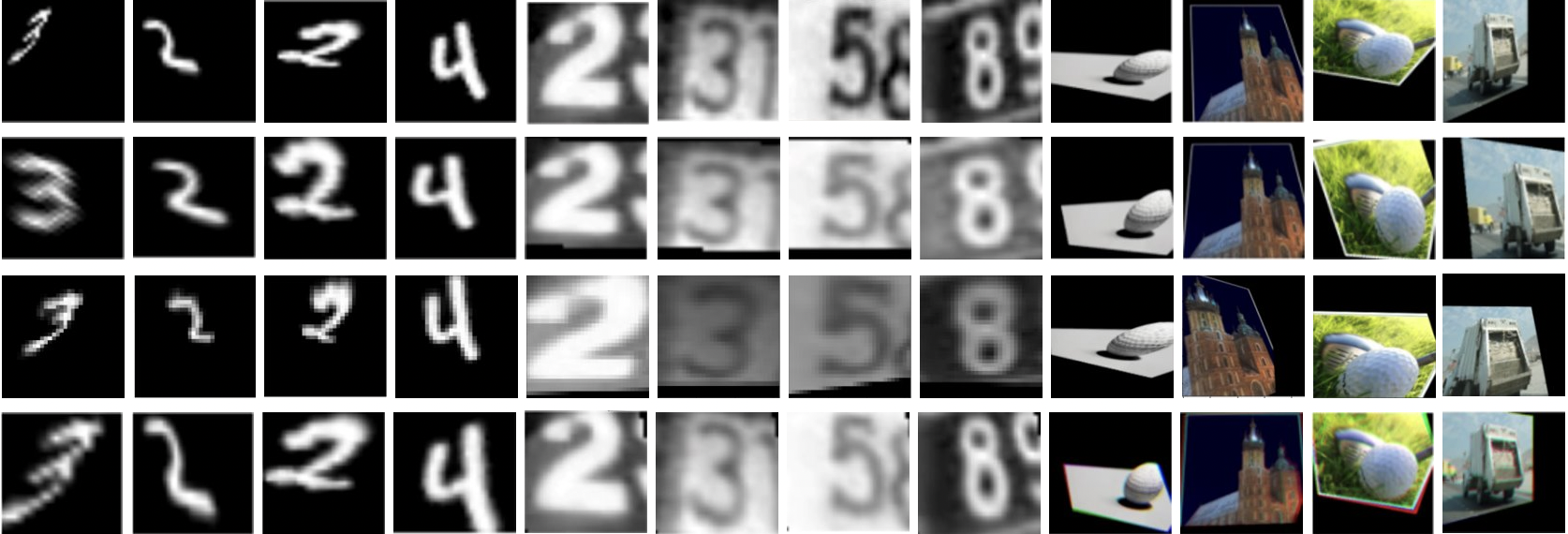}
    \vspace{-0.2cm}
    \caption{Visualizations of the layer or module outputs. Top row: input images; Second row: outputs of the STN; Third row: outputs of the ETN; Last row: outputs of the PT layer.}
    \label{fig:visualization}
    \vspace{-1.5em}
\end{figure*}

\noindent \textbf{Results.}
Tables~\ref{table:compare1} and~\ref{table:compare2} illustrate the superiority of the proposed PT layer.
With different models on the original and distorted datasets, the proposed PT layer always has better classification accuracies than both the baseline model and the models with ETN and STN modules.
The PT layer significantly improves the accuracy, especially when it is difficult for a baseline model to handle a dataset. 
For example, when we use the VGG-16 for the distorted and original Imagenette, the PT layer can provide approximately $10\%$ more accuracy than the baseline. 
% We could also observe that training using ResNet-18 architecture(without pretrained weights) improves the performance on original Imagenette by approximately $5\%$.

We also visualize some after-layer and after-module outputs from the PT layer, the STN, and the ETN. Some example input images and their corresponding outputs of the PT layer and the STN and ETN modules are shown in Figure~\ref{fig:visualization}. 
% The STN produces a single affine transformation for each image, and the ETN tries to decompose the affine transformation and has heavier impacts on the pixel intensities. 
% On the other hand, to facilitate the downstream classification, 
Compared to the STN and the ETN, the proposed PT layer learns good viewpoints (we are just displaying one of them) without cropping out the objects, highlights (centralizes) the target objects, and has smaller changes to the overall pixel intensities.

\vspace{-1.0em}
\section{Conclusion}
\label{sec:conclusion}

This paper introduces a PT layer that learns viewpoints on images in deep learning. 
The proposed PT layer learns homography to reflect the changes in the 3D world to a 2D image, provides the capacity of learning multiple and adjustable viewpoints, and facilitates its training of TMs with gradient descent by developing differentiable operations.
Experimentally, we have validated the basic functionality of the proposed PT layer, confirmed the PT layer's promising performance by the ablation study, analyzed the increased learning capacity of multiple TMs, and demonstrated our superiority as compared to other state-of-the-art methods.
% At the next step, we will explore more application scenarios in 3-dimensional image processing.

% \section*{Acknowledgement}

% The funding information is omitted for the peer review process.
% \clearpage

\bibliographystyle{IEEEtran}
\bibliography{bibliography}

\end{document}